\documentclass{sig-alternate}
\usepackage{url}
\usepackage[latin1]{inputenc}
\usepackage[usenames,dvipsnames]{color}
\usepackage{color}%
\usepackage{amsmath}%
\usepackage{bm}%

\newcommand {\out}[1]{}

\begin{document}

\conferenceinfo{USEWOD}{2012 Lyon, France}

\title{Learning to Rank Query Recommendations\\ by Semantic Similarity}

\numberofauthors{3}

\author{
  \alignauthor Sumio Fujita\\
  \affaddr{Yahoo! JAPAN Research}\\
  \affaddr{Midtown Tower, Akasaka}\\
  \affaddr{Tokyo 107-6211, Japan}\\
  \email{sufujita@yahoo-corp.jp}
  \alignauthor Georges Dupret\\
  \affaddr{Yahoo! Labs}\\
  \affaddr{701 First Avenue, Sunnyvale}\\
  \affaddr{CA, 94089-0703, USA}\\
  \email{gdupret@yahoo-inc.com}
  \alignauthor Ricardo Baeza-Yates\\
  \affaddr{Yahoo! Research}\\
  \affaddr{Diagonal 177, 9th floor}\\
  \affaddr{Barcelona, Spain}\\
  \email{rbaeza@acm.org} }

\maketitle

\begin{abstract}
  The web logs of the interactions of people with a search engine show that users often
  reformulate their queries. Examining these reformulations shows that
  recommendations that precise the focus of a query are helpful,
  like those based on expansions of the original queries. But it also
  shows that queries that express some topical shift with respect to
  the original query can help user access more rapidly the information
  they need.

  We propose a method to identify from search engine query logs possible
candidate queries that can be recommended to focus or shift a topic. This
  method combines various click-based, topic-based and session based
  ranking strategies and uses supervised learning in order to
  maximize the semantic similarity between the query and the
  recommendations, while at the same time we diversify them.

  We evaluate our method using the query/click logs of a Japanese web
  search engine and we show that the combination of the three methods
  proposed is significantly better than any of them taken
  individually.
\end{abstract}


\pagebreak

\section{Introduction}
\vspace{-2mm}%
\subsubsection*{Problem Statement}
Information retrieval is often an interactive process where a user
successively refines his or her original search query, switch focus
and approach her/his goal in several steps. Assisting users in this
process makes it less cumbersome. Query suggestions are particularly
useful on mobile devices and for Asian languages with complex
character sets where typing queries is particularly inconvenient and
time consuming.

Query recommendation engines should not limit themselves to proposing
more focused queries, but should also suggest\out{surface} queries
that are a reasonable switch in focus. This is confirmed by examining
search engine query log data. For example, the most frequent queries
after ``toyota'' are ``honda'', ``nissan'' and ``lexus'', none of
which are a direct refinement of the original query.  As another
example, the most frequent query after ``driver's license renewal'' is
``slight violence of traffic laws'', which may prevent drivers from
renewing their driver's license.

Search engines sometimes suggest queries with some additional
modifiers, focusing on a particular aspect of the previous query.
According to Jansen {\em et al.}~\cite{jansen09jasist}, queries which
initiated a new session are in 31\% cases followed by query
reformulations of the type `specialization' or `specialization with
reformulation'.  Such drill down operations are not necessarily
observed more frequently than topic shifting. Topic shifting occurs
especially when users engage in complex tasks like researching for a new
vehicle and comparing competing candidate models, or when they look
for information on how to renew a driver license including ancillary
tasks, like discovering office hours, finding the required forms, the
office address, etc. Boldi~\cite{boldi2009wiiat_br} pointed out that the
typically useful recommendations are either specializations or topic
shifting, which they refer to as ``parallel moves''.

Unlike pre-retrieval query suggestions, which frequently propose
automatic query completion right in the query box, query
recommendation provides semantically related queries and exclude
trivially synonymous queries, since state-of-the-art commercial search
engines are good enough to cover minor spelling variations or even
some miss-spellings. Nevertheless, diversifying query recommendations
 would help for polysemic queries.

\subsubsection*{Methodology}
Query recommendations are often based on clustering methods with the
inconvenience that queries falling in the same cluster are some time
more ambiguous and less helpful than the original query.  Instead, we
formulate in this work three distinct methods of extracting query
recommendations from a search engine's click-through logs.  These
methods induce directed links between queries existing in the logs and
hence have the potential to overcome the limitations of the clustering
methods.  The \textit{first} method is based on the position of the
clicked URLs in the search ranking of the original query and its
potential recommendations.  The \textit{second} is based on
reformulations of the original query that can be easily detected in
the logs using the query surface forms.  Users reformulate queries for
a variety of reasons: because the original formulation is too
ambiguous or carry other meanings they did not intend, or because the
results returned by the engine are not adequate.  The \textit{third}
method is also based on query reformulations but it is based on
co-occurrence relations of the queries\out{ and URLs} in the sessions.
We show that each method has its own advantages and drawbacks.  The
first method sometimes leads to recommendations that are more
difficult to understand because it tends to include Web jargon, but it
is sometimes more useful than the simple reformulation method because
it leverages the topical knowledge of other users.  By construction, the
second method rarely drifts from the original search topic and tends to
be limited to specializations of the original query. This results in
safer recommendations with less coverage.  Variants of this method are
used by many commercial search engines because it is safer and more
predictable.  The last method is better suited for shifting topics
because it provides more diverse recommendations such as
\textit{parallel move} reformulations~\cite{boldi2009wiiat_br}.  On the
other hand, trivial variants or completely unrelated queries are not
useful. Each method has distinct capabilities and short-comings, and then
it would be interesting to develop a method that chooses the best
candidates and offer to the user an improved set of
recommendations. This is the objective of this work.

\subsubsection*{Assumptions}
\begin{figure}[ht]
      \begin{center}
        \includegraphics*[width=58mm] {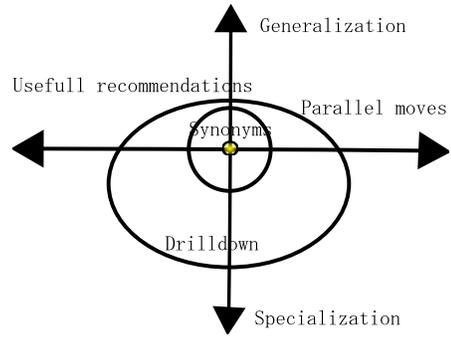}
        \caption{Schematic view of semantic relations of related queries.}
        \label{fig:query_relations}
      \end{center}
\end{figure}

Since most useful recommendations are either specializations or
parallel moves, it is better to use distinct methods to cover both
types.  It is also necessary to exclude trivial synonyms and unrelated
queries.  We make the following assumption: in the semantic hierarchy
of information needs, locating the original user query at the center,
generalization queries reside in the upper part of the hierarchy and
specialization queries in the lower. The neighbouring queries in the
semantic hierarchy are generally useful. In order to identify such
queries, we combine the recommendation candidates from three methods
and learn the ranking function according to semantic similarities
reflected in the topological relations in the semantic hierarchy.  
We schematized the relations in Figure~\ref{fig:query_relations},
where too close queries are not useful as recommendations.
On the other hand, either specializations or parallel moves are useful
to help the searcher with drill down or shift operations respectively.

\subsubsection*{Contribution}
The problem we address in this paper is how to combine such candidate
recommendations with different characteristics to diversify them.  One
possible solution is to infer the intention of the user: does she/he
intend to drill down into the topic or will she/he quit the current
sub topic and move to the next sub topic? This would undoubtedly be a
very hard task.  A priori, any query may be followed by the user
drilling down for more precise information or shifting the
intention. This depends among other things on the quality of the
results the user finds on the result page. User decisions and
consecutive search actions are not only query dependent but also user
and context dependent.  Instead of attempting to predict the user's
state of mind, we propose to minimize the risk of dissatisfying the
user by proposing carefully various solutions.  
Expressed in the terms of our prior assumption, we try to
maximize the semantic similarity between the original query and
recommended queries by combining different types of recommendations,
to make them more diverse.  We will not use lexical features such as
semantic categories of query terms, since such lexical knowledge has a
usually fairly limited coverage.

\subsubsection*{Organization}
In Section~\ref{related_work}, we present some related works that make
use of query and click logs.  We present the methods used to extract
the different types of recommendations in Section~\ref{relations}.  We
show empirically that the session based method is good at identifying
shifting queries whereas the other two methods favor focused faceted
queries.  We combine these three methods to maximize the semantic
similarity measure in Section~\ref{query_similarity}.  We describe the
supervised learning algorithm we use in Section~\ref{combining}.  In
Section~\ref{experiments}, we report the results of an empirical
study based on the click logs of a popular Japanese search
engine and Section~\ref{conclusions} concludes the paper.

\section{Related Work}
\label{related_work}
\subsubsection*{Click Log Analysis}
Click logs typically contain information such as the search query
string, time stamp, browser identifier, clicked URLs, and rank
positions.  Although correctly interpreting clicks is not
straightforward~\cite{joachims2005aic,dupret2010wsdm},
click information is often used as an implicit feedback on URL
relevance.

Beeferman and Berger studied Web search query logs and clustered
click-through data by iterating two steps: (a) combining the two most
similar queries and (b) combining the two most similar
URLs~\cite{beeferman2000acs}. The generated clusters were used to
enhance query recommendations. Baeza-Yates {\em et al.}\ proposed a
query recommendation technique using query clustering based on the
similarity of clicked URLs~\cite{baeza2007jasist}. Dupret and Mendoza
also addressed query recommendation using click-through data but
focused on document ranking~\cite{dupret2005spire}. Xue {\em et
  al.}~\cite{xue2004} used click-through data to create metadata for
Web pages. They estimated document-to-document similarities on the
basis of co-clicked queries and query strings used as metadata or
tags. These estimates were then spread over similar
documents. Craswell and Szummer, who used click-through data for image
retrieval, experimented with backward random
walks~\cite{craswell2007rwc}. Their method is based on
query-to-document transition probabilities on a click
graph. Baeza-Yates and Tiberi extracted semantic relations between
queries on the basis of set relations of clicked
URLs~\cite{baeza2007}.  \out{Li {\em et al.} used a set of queries as a
seed set and propagated these labels through the click graph. They
applied this method to semi-supervised query classifications of two
search intent classes, namely product intent and job
intent~\cite{li2008lqi}.  Fujita {\em et al.} applied topic sensitive
Pagerank like score propagation to the click bipartite
graphs~\cite{fujita2010riao}. By expanding the graph using URL
structure, they improved the effectiveness of query attribute
estimation.  These two methods use a random-walk-based soft clustering
approach to identify similar queries. They are adequate for
applications like estimating query attributes but they suffer the same
problem as other clustering methods when applied to the query
recommendation task.}  Antonellis {\em et al.} proposed the
Simrank++~\cite{antonellis08vldb} method in which query similarity is
propagated through click bipartite graphs.  They used the query
similarity measure to rewrite queries in order to extend advertisement
matching.  Again, such a measure of structural-context similarity
might be adequate for the task such as query rewrites for sponsored
search where rewriting to a practical synonymous query is effective.

\subsubsection*{Term Expansion Based}
\out{ Anick and Tipirneni proposed ``lexical dispersion hypothesis''
by which they induced terminological feedback from search results
documents~\cite{anick99sigir}. Anick evaluated the terminological
feedback for the web search assistance by examining the logs of search
user activities~\cite{anick03sigir}.Their method extracts query terms
from the result documents but we adopted similar ideas to queries in
the logs.}  Jones {\em et al.} extracted query substitutions from the
same user sessions by identifying correlated term pairs and
substituting phrases~\cite{jones06www}.  Jones' work addressed query
rewrites in sponsored search contexts where the ``precise rewriting''
such as ``{\em automobile insurance} $\mapsto$ {\em automotive
insurance}'', is mostly preferred.  However, the current state of the
art search engines return very similar results to these two queries.

\subsubsection*{Query Session Based}
Spink {\em et al.} surveyed information related to successive Web
searches~\cite{spink98ir} and found that the information involves
changes and shifts in search terms, search strategies, and relevance
judgments. Jansen {\em et al.} analyzed successive queries in large
Web search query logs~\cite{jansen09jasist}, and He {\em et al.} tried to
detect session boundaries on the basis of search patterns and time
intervals in query logs~\cite{he02ipm}.  Fonseca {\em et al.}
extracted query relations by using association rules from the same
user sessions~\cite{fonseca05cikm}. Boldi {\em et al.}  analyzed
search user sessions, classified query reformulation
types~\cite{boldi2009wiiat_br}, and derived query-flow graphs for the
extracted query recommendations. They pointed out
that the typically useful recommendations are either specializations
or parallel moves while trivial variants or completely unrelated queries
are not useful.\out{Bordino {\em et al.} evaluated query similarity
measures on query-flow graphs with and without graph
projections~\cite{bordino2010}.} Cao {\em et al.}  applied click-based
clustering to session-based query suggestions~\cite{cao08kdd} and they
claim that the {\em context awareness} helps to better understand
user's search intent and to make more meaningful suggestions.  However, 
they do not evaluate well if the {\em context awareness} really improves
the suggestion utility due to the lack of an adequate baseline.

\section{Generating Candidates}
\label{relations} 

\out{
In this work we focus on the generation of query recommendations
through the use of inter-query relations in Web search logs. The
queries observed in the logs have mainly been reformulated or
rewritten on the basis of the search results. The method we proposed
here identifies the inter-query relations and uses them to recommend
candidate queries. These recommendations represent either a
specialization of the information need, a refinement of the original
query, or a parallel move from the original search intent. Different
methods are required to extract these different types of query
reformulation from the logs.

Out of the three methods, only the first uses click information. The
other two simply extract from the query logs syntactically expanded 
multi-faceted queries that were entered by a certain number of
users.  One method considers co-topic relations while the other
considers co-session relations. }

In this work we focus on
the generation of query recommendations through the use of inter-query
relations in Web search logs.  As we have seen in the previous section,
log based query recommendation techniques fall into one of three
approaches, namely click-based, term expansion based and session based.
Each approach intends to capture patterns of different user activities
from the query logs.  Queries are related by a co-click relation in view of
users clicking on the same URL in response to them.  Queries are also
related to their possible expansion by adding facet modifiers, {\em
i.e.} co-topic relations. Finally, queries are related by their co-occurrence
in a user session.

The following three methods extract these three types of
inter-query relations representing user behaviors in the logs: either
a specialization/refinement of the information need or a parallel move
from the original search intent.  The methods are simple although
they are intended to extract candidates thoroughly, so
that they are adequately combined and re-ranked by a supervised
learning algorithm to maximize the semantic similarity measure.

\subsection{Best Rank Directed Co-Click Relations}
\label{best_rank_directed}

\newcommand{\argmin}{\mathop{\rm arg~min}\limits}
This method compares the positions in the search results of the
documents clicked during a query session. If a query $q'$ different from
query $q$ better orders ---according to a suitable measure---the
clicked documents in a significant number of sessions of $q$, then
$q'$ is a candidate query for recommendation. There is a fundamental
basis for considering the clicked document rankings rather than the
simple similarity of clicked page sets. Take for example the
multi-faceted query ``curry,'' The documents that a user selects can
help identify a posteriori his information need: if he or she is
interested in how to cook curry, he or she will select pages related
to cooking rather than those related to the origin of ``curry'' in
Indian culinary history. The assumption is that savvier users with the
same information need will probably express the query less ambiguously
and enter ``curry recipe,'' for example, as the query. The hypothesis
we wish to investigate here is whether documents clicked by a previous
user are ranked higher in the ``curry recipe'' results than in the
``curry'' results. If they are, we can retrieve the ``curry recipe''
query from the log and recommend it.

More formally, suppose that $u$ is a clicked URL\footnote{We use
  ``document,'' ``page,'' and ``URL'' interchangeably.} in the results
for query $q$. For each such clicked URL $u$, we assume the existence
of a set of queries for which URL $u$ is ranked higher in the
results. This set might be empty. We hypothesize that such queries are
potential recommendations for~$q$.

\out{
We define the URL cover $UC_{q}$ of a query~$q$ as the set of URLs
clicked in response to query $q$, and the query cover of URL~$u$,
$QC_{u}$ as the set of queries for which URL~$u$ is clicked.}

\out{
The co-click query set, $CCQ_{q,u}$, the set of queries in response to which URL
$u$ in $UC_{q}$ is also clicked, is defined as
\begin{eqnarray*}
  CCQ_{q,u} &\equiv& \{q'|q' \in QC_{u}, u \in UC_{q}\}~.
\end{eqnarray*}}

\out{The best rank $BR_{u}$ of URL $u$ is the minimum (best) rank this URL
achieves in its query cover set. We first define $\mbox{rank}_q(u)$ as
the rank position of URL $u$ for query $q$. We then have
\begin{eqnarray*}
  BR_{u} &\equiv& \min_{q \in QC_{u}} \mbox{rank}_q(u)~.
\end{eqnarray*}}

We first define the URL cover $UC_{q}$ of a query~$q$ as the set of
URLs clicked in response to query $q$, and the query cover of URL~$u$,
$QC_{u}$ as the set of queries for which URL~$u$ is clicked.\out{  The
co-click query set, $CCQ_{q}$, the set of queries in response to
which at least a URL $u$ in $UC_{q}$ is also clicked, is defined as
\begin{eqnarray*}
  CCQ_{q} &\equiv& \{q'|\exists u: u \in UC_{q} \wedge u \in UC_{q'}\}~.
\end{eqnarray*}}
 We define $\mbox{rank}_u(q)$ as the rank position of URL $u$ for
query $q$.  The set of best rank co-click queries for query $q$,
$BRCCQ_{q}$, is as follows:
\begin{eqnarray*}
  BRCCQ_{q} &\equiv& \bigcup_{u \in UC_{q}} \argmin_{q' \in QC_{u}} \mbox{rank}_{u}(q')~.
\end{eqnarray*}

\out{
\begin{eqnarray*}
  BRCCQ_{q,u} &\equiv& \{q'|q' \in QC_{u}, \mbox{rank}_{q'}(u) = BR_{u}\}~.
\end{eqnarray*}}

We estimate the strength of the relations between a query and its
candidate recommendations in accordance with the following weighting
scheme. We define~$cnt(u,q)$ as the number of clicks on~$u$ in
response to query~$q$,~$cnt(q)$ as the total number of clicks in
response to query~$q$,~$cnt(u)~$ as the total number of clicks on~$u$
regardless of the query and~$Q$ as the set of all queries. We define
the probability~$P_{CC}(q_2|q_1)$ as follows:
\newcommand{\pcc}{P_{CC}}
\begin{eqnarray*}
  \pcc(q_2|q_1) &=& \sum_{u \in UC_{q1}} P(u|q_1) \cdot P(q_2|u) \\
  &=& \sum_{u \in UC_{q1}} P(u|q_1) \cdot \frac{P(q_2)\cdot P(u|q_2)}{P(u)} 
\end{eqnarray*}
with 
\begin{eqnarray*}
  P(u) &=& \frac{cnt(u)}{\sum_{q \in Q} cnt(q)}~, \\
  P(q) &=&  \frac{cnt(q)}{\sum_{q' \in Q} cnt(q')}~,~{\rm and}\\
  P(u|q) &=& \frac{cnt(u,q)}{cnt(q)}~.
\end{eqnarray*}

This approach can be regarded as a special case of the session-based
recommendation proposed by Dupret and Mendoza
\cite{dupret2005spire}. In this approach, each single click is
considered to be a single session. This is clearly distinct from the
approach used in \textit{query clustering methods} because it
explicitly uses the positions of the documents in the results list.
\out{
One issue with this method is its coverage. If
}

\subsection{Co-topic Relations}

Commercial search engines commonly use expansions of input query
string in logs as recommendations. Here, we introduce a variation that
takes advantage of a characteristic of the Japanese language. \out{As
mentioned in Fujita {\it et al.}~\cite{fujita2010riao}, the}The agglutinant
nature of the Japanese language makes it comparatively easy to detect
topic-facet structure in queries.  In practice, a facet directive in
Japanese is easily identified as a word that appears as the last term
of a significant number of distinct queries. In our experiments, if a
word is the last of at least five distinct query strings, it can be
safely regarded as a facet word as long as queries appearing fewer
than ten times are eliminated from the logs. Thus, from the
topic-part-only query ``curry'', we may induce ``curry recipe'',
``curry restaurant'', and other queries with different directives.

We define a co-topic query as a query expanded by the addition of a
facet directive.  As for co-click relations, we define a weighting
scheme that captures the strength of the relation between the original
query $q_1$ and a co-topic recommendation~$q_2$ based on the following
probability: we first define $CTQ_{q_1}$ as the set of co-topic
queries formed over $q_1$, the similarity is expressed as:
\newcommand{\pct}{P_{CT}}
\begin{eqnarray*}
  \pct(q_2|q_1) &=& \frac{cnt(q_2)}
  {cnt(q_1) + \sum_{q_{2'} \in CTQ_{q_1}} cnt(q_{2'})}~.
\end{eqnarray*}

This relation normally represents a specialization of the original
concept by adding a facet directive which restrictively modifies the
original concept.
\out{
The intuition behind this method is closely related with the notion of
``lexical dispersion'' discussed by Anick and
Tipirneni~\cite{anick99sigir} who propose to extract compound query
families from search result documents. The method proposed here is
simpler but turns out to be very effective in practice.}

\subsection{Co-Session Relations}
\label{co-session_relations}

This last method identifies the query reformulations observed a
significant number of times during the sessions of users. Co-session
queries are queries submitted consecutively from the same user in a
time interval typically no longer than 5 minutes. Co-session queries
includes not only the reformulation or rewriting of queries, such as
in the co-topic relation, but also queries that reflect a shift in
information needs. (A more complete nomenclature of the relations
extracted this way can be found in~\cite{boldi2009wiiat_br}.)

We define the set $CSQ_{q_1}$ as the set of queries sharing a
co-session relation with~$q_1$. The strength of a co-session relation
between $q_1$ and $q_2$ is estimated as a probability:
\newcommand{\pcs}{P_{CS}}
\begin{eqnarray*}
  \pcs(q_2|q_1) &=& \frac{cnt(q_2,q_1)}{cnt(q_1)}~,
  \out{ \pcs(q_2,q_1) &=& \frac{cnt(q_2)}
    {cnt(q_1) + \sum_{q_{2'} \in CSQ_{q_1}} cnt(q_{2'})}~.}
\end{eqnarray*}
where $cnt(q_2,q_1)$ denotes the count of the query $q_2$ preceded by
the query $q_1$ in the same user session.

This method is relatively robust to mistakes during the
segmentation of user activities in session: if $q_2$ and $q_1$ do not
belong to the same session, $cnt(q_2,q_1)$ will be small, leading to a
relation with a low strength.

\section{Query Similarity}
\label{query_similarity}

It is not straightforward to assess the quality of query
recommendations. To evaluate the three methods presented in the
previous section, we use the semantic similarity of the queries after
they are mapped into a category hierarchy.\out{ We first define a
similarity measure between query pairs in
Section~\ref{category_similarity} and different diversity measures
over set of queries in Section~\ref{sec:diversity_measure}. We then
proceed to the evaluation of the three methods.

\subsection{Category Similarity}
\label{category_similarity}

\sloppy
} We adopt a similarity measure between query pairs by 
Baeza-Yates and Tiberi~\cite{baeza2007} who evaluated semantic relations
between queries connected by an edge of their click cover graph. For this
purpose, they use the Open Directory
Project\footnote{http://www.dmoz.org/}, where queries are
matched against the directory content to find the categories where they
belong. We apply the same methodology but using the Yahoo!
JAPAN directory\footnote{http://dir.yahoo.co.jp/} because it
has a more complete coverage of Japanese queries.  

Baeza-Yates and Tiberi use the following similarity function on the
categories matching two queries $q$ and $q'$:
\begin{eqnarray*}
  Sim_{prefix}(D,D') =
  |P(D,D')|/max\{|D|,|D'|\}~,
\end{eqnarray*}
where $P(D,D')$ is the longest common prefix of the category paths $D$ and
$D'$ where the queries $q$ and $q'$ were found, respectively. 
This is intuitively reasonable: consider for example the query
``Spain''. The query term is found in ``Regional / Countries / Spain''
while ``Barcelona'' is found in ``Regional / Countries / Spain /
Autonomous Communities / Catalonia / Cities / Barcelona,''. Then, the
similarity is~$\frac{3}{7}$.  

However, we needed to make some adjustment because in the Yahoo!
directory, a subcategory like ``Spain'' might appear below
diverse top categories such as ``Maps / By region / Countries'',
``Arts / By region / Countries'', or ``Recreation / Travel / By region
/ Countries''. We therefore use the following similarity function:
\begin{eqnarray*}
  Sim_{substring}(D,D')&=&\frac{C(D,D')}{max\{|D|,|D'|\}}~,
\end{eqnarray*}
where $C(D,D')$ is the number of common subparts of two category
paths that match the queries. The previous similarity function measures the ratio of the
hyper concepts that the two categories share whereas this new function
considers the facet similarity of subcategories.

\sloppy To associate Yahoo! categories with each query, we used the
directory search application programming interface (API), which
returns a list of categorized sites retrieved by ``AND'' boolean
queries. This presumably favors co-topic relations over co-click
relations because registered sites retrieved by the expanded query $q_2$
are also retrieved by original query $q_1$ due to the ``AND''
operation.  As a categorized site is retrieved, the procedure votes to
its category.  The category with maximum number of votes is assigned
to the query.  Inter query similarity depends on similarities of
category pairs, and the maximum similarity through category pairs was
selected as the final score.

 For query recommendation, queries that are virtually the same are
useless, so we excluded queries falling in the group of trivial variants.
Queries were grouped in accordance with the clicked URL set $UC_q$ by
an online single-pass clustering using a vectorial representation of
each URL set, where the component is the click frequency of the URL in
response to the query.

\section{Combining Recommendations}
\label{combining}

Identifying the user intention from contextual information is a very
difficult task and is not guaranteed to be effective. Instead, we take
a more conservative approach and we combine the three methods
described above. We attempt to take advantage of each method strength
but also hedge against bad recommendations by providing some
conservative specialization queries, some serendipitous queries and by
proposing some ``topic shifting'' queries. In other words, we
diversify the set of recommended queries.


We formulate the problem as a ``learning to rank'' task for which we
use the similarity measure defined in
Section~\ref{query_similarity}. We use gradient boosting decision
trees~(GBDT) described in~\cite{friedman00annals} because of the
robustness to overfitting, the scalability and the ability to handle
highly non-linear problems of this method.

\out{
\sloppy
Following the assumption that the neighbouring items in the category
hierarchy are most useful as recommendations~(see
Fig.~\ref{fig:query_relations} of the introduction), we try to combine 
three ranking lists of candidate recommendations obtained by the different
methods described above. The criteria is the maximization of the
similarity between the original query and its recommendation.  We
approach this task as a ``learning to rank'' problem for which we
adopt the {\em gradient boosting decision tree} (GBDT)
method~\cite{friedman00annals}.
}
\out{
Adopting the similarity of pairs as target attribute, the learned
model tries to rank the most similar queries, which might be trivial
variants, at the highest positions.  To avoid this, we eliminated
recommendations belonging to the same query cluster as the original
based on the query clustering described in~\ref{query_similarity}.}

\subsubsection*{Training Data}
For training and test pairs, we calculated the similarity measure
described in Section~\ref{query_similarity} as the target
attribute. \out{ Not only the recommended pairs, we prepared random
query pairs to augment the negative data.} For this we cleaned the
data and added random query pairs to augment the number of negative
examples and balance the training set. The details are given in
Section~\ref{experiments}.

\subsubsection*{Feature Set}

We defined the quantities $\pcc(q_2|q_1)$, $\pct(q_2|q_1)$ and
$\pcs(q_2|q_1)$ in Section~\ref{relations}.  On top of these features,
we defined 24 features as described in Table~\ref{tab:features}.  {\em
Facet extraction features} are extracted from the query logs.  We
adopted the {\em query textual features} used
in~\cite{boldi2009wiiat_br}.  Cosine similarities are computed based on
the bag of character bigrams and on chunks, i.e. contiguous character
strings split by a white space. As {\em result click features}, we
measure how the queries are multi-faceted with respect to user
behavior on the result sets. Click entropy is used to reflect query
ambiguity as in Teevan~{\it et al.}~\cite{teevan08sigir}.  For {\em
query session co-occurrence} we derive features from pair of
queries directly following one another in a user session. $LLR$ is
adopted from Jones {\em et al.}~\cite{jones06www}. We introduced this to
identify significant query pairs from sessions. A high value means a
strong dependency between two adjacent queries in a session.

\begin{table}[ht]
  \caption{Features used for the supervised learning.}
  \label{tab:features}
  \begin{center}
    \begin{tabular}{l|p{2in}}
      \hline
      \multicolumn{2}{c}{Facet extraction features}\\
      \hline\hline
      $\pcc(q_2|q_1)$
      & co-click query probability\\
      $\pct(q_2|q_1)$
      & co-topic query probability\\
      $\pcs(q_2|q_1)$
      & co-session query probability\\
      $Freq.q1$
      & Click frequency of $q_1$\\
      $Freq.q2$
      & Click frequency of $q_2$\\
      $Freq.topic$
      & Total topic frequency of $q_1$\\
      \hline
      \multicolumn{2}{c}{Query textual features}\\
      \hline\hline
      $Len.q_1$
      & Character length of $q_1$\\
      $Len.q_2$
      & Character length of $q_2$\\
      $CLen.q_1$
      & Chunk length of $q_1$\\
      $CLen.q_2$
      & Chunk length of $q_2$\\
      $delta.Len$
      & $Len.q_2-Len.q_1$\\
      $delta.Len.Rel$
      & $(Len.q_2-Len.q_1)/Len.q_1$\\
      $delta.CLen$
      & $CLen.q_2-CLen.q_1$\\
      $delta.CLen.Rel$
      & $(CLen.q_2-CLen.q_1)/CLen.q_1$\\
      $mb.Leven$
      & Levenshtein distance of $q_1$ and $q_2$ by multi-byte character basis\\
      $Leven$
      & Levenshtein distance of $q_1$ and $q_2$ by single-byte basis\\
      $CCos$
      & Cosine similarities between bag of chunk (keyword) representations of 
      $q_1$ and $q_2$\\
      $BCos$
      & Cosine similarities between bag of character bigrams representations of 
      $q_1$ and $q_2$\\
      \hline
      \multicolumn{2}{c}{Result click features}\\
      \hline\hline
      $Ent.q_1$
      & Search result click entropy of $q_1$\\
      $Ent.q_2$
      & Search result click entropy of $q_2$\\
      $delta.Ent$
      & $Ent.q_1-Ent.q_2$\\
      \hline
      \multicolumn{2}{c}{Query session co-occurrence features}\\
      \hline\hline
      $Next.Ent$
      & Entropy of the query following $q_1$\\
      $LLR$  
      & Log likelihood ratio of observing $q_2$ after $q_1$ in the same session\\
      \hline
      \multicolumn{2}{c}{Target attribute feature}\\
      \hline\hline  
      $Sim$
      & Category similarity between $q_1$ and $q_2$\\
      \hline
    \end{tabular}
  \end{center}
  \vspace*{-2.5mm}
\end{table}

\subsubsection*{Learning Models}

As mentioned above, we use gradient boosting decision
trees~(GBDT~\cite{friedman00annals}). This is an additive regression
model over an ensemble of shallow regression trees.

It iteratively fits an additive model:
\begin{eqnarray*}
  F_{m}(x) &=& F_{m-1}(x)+\beta_{m}T_m(x;\Theta_m)~,
\end{eqnarray*}
where $T_m(x;\Theta_m)$ is a regression tree at iteration $m$,
weighted by parameter $\beta_m$, with a finite number of parameter
$\Theta_m$, consisting of split regions and corresponding weights,
which are optimized such that a certain loss function is minimized as
follows:
\begin{eqnarray*}
  (\beta_m,\Theta_m) &=& argmin_{\beta,\Theta}
  \sum_{i=1}^N L(y_i,F_{m-1}(x)+\beta T_m(x;\Theta))~.
\end{eqnarray*}

At iteration $m$, tree $T_m(x;\Theta_m)$ is induced to fit the
negative gradient by least squares:
\begin{eqnarray*}
  \hat \Theta &=& argmin_{\Theta,\beta}
  \sum_{i=1}^N \left(-g_m(x_i)-\beta_{m}T_m(x;\Theta_m)\right)^2~.
\end{eqnarray*}
where $-g_m(x_i)$ is the gradient over current prediction function:
\begin{eqnarray*}
  -g_m(x_i) &=& - \left[ \frac{\partial L(y_i,F(x_i))}{\partial F(x_i)}\right]_{F(x)=F_{m-1}(x)}~.
\end{eqnarray*}

Each non-terminal node in the tree represents the condition of a split
on a feature space and each terminal node represents a region. The
improvement criterion to evaluate splits of a current terminal region
$R$ into two subregions $(R_{\ell},R_r)$ is as follows:
\begin{eqnarray*}
  i^2(R_{\ell},R_r) &=& \frac{w_{\ell}w_r}{w_{\ell}+w_r}(y_{\ell}-y_r)^2~,
\end{eqnarray*}
where $y_{\ell}$ and $y_r$ are the mean response of left and right subregions,
respectively, and $w_{\ell}$ and $w_r$ are the corresponding sums of weights.
We evaluate the relative importance of each feature by the normalized
sum of $i^2(R_{\ell},R_r)$ through all the nodes corresponding to the
feature.
\out{
As the target of the learning, we considered the strategy to maximize
the category similarities following the assumption that the
neighbouring items in the category hierarchy are most useful as
recommendations.
}
\out{
\subsection{Maximizing Category Similarities}
\label{}
In this strategy, the target attribute is the category similarity
measure between the paired queries. The training and test data are
easily obtained from the extracted query pairs.
}

\section{Experiments}
\label{experiments}

\subsection{Evaluation Data}
\label{evaluation_data}

To evaluate our proposed combined method, we used a sample of the
query log of a Japanese commercial search engine.  First,
query-clicked URL pairs that appear only once were removed.  Second,
identical query-URL pairs with the same browser cookie (i.e., queries
from the same client) were counted only once to improve robustness
against spam. Third, we selected from the log the 4,544 queries that
contain one of the seven most frequent facet directives appearing in
Japanese web search\out{~\cite{fujita2010riao}}.  Table~\ref{tab:data}
shows the statistics of our evaluation data.  On the basis of these
initial queries, we extracted 188,737 query pairs, among which, 70,041
pairs are in a best rank co-click relation, 77,991 pairs in a co-topic
relation, and 66,612 pairs in a co-session relation.  From them, we
excluded pairs where either query failed to be assigned to any
category. At the end, we obtained 86,544 query URL pairs, which we
split into two sets to carry out a two fold cross validation.  We
supplemented training pairs by 82,212 randomly combined pairs of
queries and recommended queries, which act as negative or
counter-examples. Notice that average semantic similarities between
pairs are high for co-click pairs.

\begin{table}[ht]
  \caption{Statistics of Evaluation data. The number of categorized pairs are
    between parentheses.}
  \label{tab:data}
  \begin{center}
    \begin{tabular}{l|c|c}
      \hline
      Data type & Numbers & Avg. sim.\\
      \hline\hline
      Original queries
      & 4,544 (--) & --\\
      Co-click pairs
      & 70,041 (25,114) & 0.8075\\
      Co-topic pairs
      & 77,991 (28,454) & 0.7954\\
      Co-session pairs
      & 66,612 (41,179) & 0.6837\\
      Combined pairs
      & 188,737 (86,544) & 0.7326\\
      Random pairs
      & 188,737 (82,212) & 0.3215\\
      \hline
    \end{tabular}
  \end{center}
  \vspace*{-2.5mm}
\end{table}

\subsection{Evaluation Measures}
\label{evaluation_measures}

Given a ranked query list $Q$, the {\em discounted cumulative gain}
(DCG) at the rank threshold $R$ is defined as follows:
\begin{eqnarray*}
  DCG_R(Q) &=& g_1 + \sum_{r=2}^R \frac{g_r}{\log_2 r}~,
\end{eqnarray*}
where $g_r$ is the score according to the judgement at the rank $r$ in
$Q$.
  
We assigned five grades to the similarity of each query pair, namely
``perfect'' (above 0.75), ``excellent'' (between 0.75 and 0.5),
``good'' (between 0.5 and 0.25), ``fair'' (below 0.25 but above 0.0),
and ``poor'' (at 0) according to the value range of similarities. We
assign scores of 10, 7, 3, 0.5 and 0.0 to these five grade labels.

The ideal ranked query list~$I$ is obtained by ranking the
recommendations in decreasing order of their label values.  It is used
to define the normalized DCG. In particular, we use the normalized DCG
at 5 (NDCG5), defined as follows:
\begin{eqnarray*}
  NDCG_5(Q,I) &=& \frac{DCG_5(Q)}{DCG_5(I)}~.
\end{eqnarray*}

The {\em average precision} (AP) of a ranked list is defined as usual:
\begin{eqnarray*}
  AP &=& \frac{\sum_{j=1}^k P(j)*R(j)}{\sum_{j=1}^k R(j)}\\
  {\rm with}~~P(j) &=& \frac{\sum_{i=1}^j R(j)}{j}
\end{eqnarray*}
where $R(j)$ is the binary judgement of the relevance of $j^{th}$ item
in the list.  We set this to 1 if the grade is ``excellent'' or better
and 0 otherwise.  The {\em mean average precision} (MAP) of a set of test
queries is the mean AP through this set.

\subsection{Ranking by a Single Method}
\label{single_method}

Table~\ref{tab:compare} compares the NDCG5 and MAP values of the
single methods and machine learned combined methods. Also included are
the results of simply taking a linear combination of the query scores
of each method computed separately.

\subsubsection*{Co-click Relations}
\label{co-click}
\out{
Table~\ref{tab:recomm_samples} shows some examples of recommendation
queries ranked by three single methods and the machine learned,
combined method.}

The BRCCQs typically represent a drill down from the original query.
It does not necessarily share any lexical part with the original query
but it shares at least a clicked document with the original query.  It
often represents specializations but sometimes parallel moves 
(``ipod'' $\mapsto$ ``itune'') or generalization (``ANA''
$\mapsto$ ``airplane'').

\subsubsection*{Co-topic Relations}
\label{co-topic}

The CTQs also represent a drill down from the original query.  It
necessarily shares some lexical part with the original query but it
does not necessarily share any clicked document with it.  As expected
from higher evaluation measures, they seem to be homogeneous because
they share the left substring. But the coverage is limited especially
for longer queries that are already specific enough. It provides
conservative recommendations but strictly limited to specialization
queries.


\subsubsection*{Co-session relations}
\label{co-session}

The CSQs might represent a drill down from the original query but it
also include topic shifts.  It does not necessarily share any lexical
part nor any clicked document with the original query.  As have been
noted, parallel move queries are characteristic of this method. For
example, against the original query ``ANA''\footnote{All Nippon
Airways or ANA offers domestic flights in Japan.}, all of the top five
recommendations are either competing traffic companies such as
``JAL'', ``Skymark'' (the names of other airline companies), JR
(railway company), or travel agent companies such as ``JTB'' and ``HIS''.
This is useful to a searcher who arranges a travel plan.  In the case
of the query: ``JR'', the names of three out of six JR regional
railway companies appear as well as ``ANA''.

\subsection{Combined Ranking}
\label{combined_ranking}

\out{From the training pairs, we induced regression trees to perform linear
regressions.}
 
We used half of the pairs for training and the rest of the pairs for
evaluation.  For training, similarity measures are used as the target
function to learn. After convergence is achieved, we use the model to
rank the queries.

Because the combined ranking uses many more features other than
$P_{cc}$, $P_{ct}$ and $P_{cs}$, the ranking is very different from a
simple mixture of three basic rankings.\out{ Again, examples of
ranking can be found in Table~\ref{tab:recomm_samples} for the three
methods and their combination.  In the case of the query ``ana'', the
combined ranking includes a good balance of results from the three
original rankings as well as for the ``jr'' case.}

As shown in Table~\ref{tab:compare}, the combined ranking learned by
GBDT achieves the best scores. The improvements from the single
methods amount to between +1.8\% and +4.1\% with NDCG5; all results
being statistically significant according to a Wilcoxon test ($p\leq
0.01$).  With MAP, the conclusions are similar.  In general, the
combination of two methods is better than any single method and
combining the three methods improves the performance further,
especially in terms of MAP.

\begin{figure}[ht]
  \begin{center}
    \includegraphics*[width=75mm] {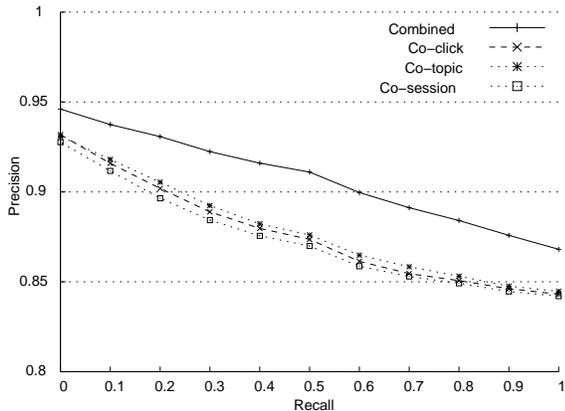}
    \caption{Precision-recall curves of co-click, co-topic, co-session
      and combined ranking of recommendations.}
    \label{fig:prec_recall1}
  \end{center}
\vspace*{-3mm}
\end{figure}

A visual inspection of Fig.~\ref{fig:prec_recall1} where the
precision-recall curves are drawn confirms these results.  The
combined ranking outperforms any single methods over the whole recall
range. As seen in the graphs, the improvement is not trivial whereas
the differences between the three single methods are small.

\begin{table}[ht]
  \caption{Recommendation ranking evaluated by NDCG5 and MAP.}
  \label{tab:compare}
  \begin{center}
    \begin{tabular}{l|r|r}
      \hline
      Ranking method
      & NDCG5 & MAP\\
      \hline
      $P_{cc}$
      & 0.9134 & 0.8570\\
      $P_{ct}$
      & 0.9238 & 0.8602\\
      $P_{cs}$
      & 0.9036 & 0.8538\\
      $P_{cc}+P_{ct}$
      & 0.9308 & 0.8716\\
      $P_{cc}+P_{cs}$
      & 0.9153 & 0.8622\\
      $P_{ct}+P_{cs}$
      & 0.9202 & 0.8660\\
      $P_{cc}+P_{ct}+P_{cs}$
      & 0.9271 & 0.8720\\
      Combined by GBDT
      & {\bf 0.9405} & {\bf 0.8978}\\
      \hline
    \end{tabular}
  \end{center}
  \vspace*{-5mm}
\end{table}

Finally, Table~\ref{tab:importance} shows the relative importance of
the features listed in Table~\ref{tab:features}.  Although $BCos$ --
the cosine similarity between the bag of character bigrams
representations of two queries -- is the most important feature
partially because of the evaluation bias mentioned in
Section~\ref{query_similarity}, other nine features account for
more than 10\% of its importance. We understand from this that the
proposed feature set is very effective for this task.  The $BCos$
feature, as well as other textual features, tends to promote queries
sharing lexical items with original, {\em i.e.}  typically found in
the CTQ sets. On the other hand, the second more important features
$LLR$ -- the log likelihood ratio of observing $q_2$ after $q_1$ in
the same session -- and $Next.Ent$ -- the next query entropy of $q_1$
-- are related to CSQs.  The $Freq.*$ features are related to the
popularity of the queries while the click entropy features $Ent.*$ are
related to the click variance. This confirms our initial hypothesis that
the three different methods of identifying potential query
recommendations are complementary and combining them is beneficial.

\begin{table}[ht]
  \caption{Relative importance of features averaging through two fold
  training sets.}
  \label{tab:importance}
  \begin{center}
    \begin{tabular}{l|l|r}
      \hline Rank & Feature & Importance\\
      \hline\hline
      1 & $BCos$ & 100.00\\
      2 & $LLR$ & 68.72\\
      3 & $P_{cc}(q_2|q_1)$ & 51.36\\
      4 & $Freq.q_2$ & 30.29\\
      5 & $Freq.topic$ & 27.80\\
      6 & $Next.Ent$ & 22.23\\
      7 & $Freq.q_1$ & 19.46\\
      8 & $mb.Leven$ & 17.26\\
      9 & $Ent.q_1$ & 17.25\\
      10 & $P_{cs}(q_2|q_1)$ & 11.91\\
      11 & $Len.q_2$ & 9.65\\
      12 & $Len.q_1$ & 7.76\\
      13 & $Ent.q_2$ & 7.63\\
      14 & $CLen.q_1$ & 6.31\\
      15 & $P_{ct}(q_2|q_1)$ & 5.02\\
      16 & $delta.Len.Rel$ & 5.01\\
      17 & $CCos$ & 2.97\\
      18 & $delta.Ent$ & 2.62\\
      19 & $delta.CLen$ & 2.03\\
      20 & $Leven$ & 1.32\\
      21 & $CLen.q_2$ & 1.30\\
      \hline
    \end{tabular}
  \end{center}
  \vspace*{-2.5mm}
\end{table}

\section{Conclusions}
\label{conclusions}

We use three methods of extracting recommendations from search logs to
improve the quality of the suggested queries. The first method
exploits the clicked document position in the ranking and selects as
candidate recommendation queries existing in the logs that have a
higher rank for the clicked document. The second method is based on
the observation that users often refine their query by adding
terms. The third method uses the query sequences in search sessions
and recommends some typical topic shifts from the query.

We carried out experiments on a sample query log of a commercial
search engine in Japan to compare the three methods. We observed that
each method has its own advantages and drawbacks: the first one, based
on the position of the clicked documents, is sometimes more difficult
to understand at first glance, but recommendations may turn out to be
more useful than those extracted from query reformulations; the second
tends to be limited to specializations of the original query, which
usually offer safer recommendations but less coverage; the last one is
good in the case of a topic shift or mission change. The preliminary
experiments conducted on the Yahoo!  directory revealed a good
semantic similarity between the extracted query pairs.  By
construction, the second method of adding a facet to a query~(CTQ)
rarely drifts from the original search topic. On the other hand, the
first method~(BRCQQ) that consists in identifying queries that would
rank higher the clicked documents tend to surface more specific,
sometimes jargon like queries. This occasionally leads to
incomprehensible recommendations, at least to our understanding
(although they might make sense for the users who issued them). CTQ
and variations on this method are used by many commercial search
engines owing to its more conservative nature but BRCCQ might be a
more effective way of recommending totally new, eye-opening queries in
a more exploratory fashion, despite the risk of recommending
over-specific or over-generic queries.  Queries extracted from user
sessions~(CSQ) provides more diverse recommendations such as
\textit{parallel move} reformulations or even topic changes if those
happen frequently in the logs ({\em e.g.} searching for an image
after having looked for some film star).

In conclusion, each recommendation method has its own merits and
drawbacks, which is the reason why we combined them. Adopting semantic
similarities as the target attribute, we learned to combine
recommendations from the three different methods in a new ranking
according to the similarity to the original query. We showed that the
resulting ranking out-performs any of the individual rankings as well
as their linear combinations in terms of NDCG5 and MAP.

\out{Discounted Cumulative Gain and
Mean Average Precision. 

As the next step of this study, we will try to select recommendations
so as to maximize the facet diversity. Consequently, we need to
evaluate the diversity in recommendation ranking. Evaluation of the
recommendation is also an important issues in this research area and
relatively less investigated than that of the document diversity.}

As the next step of this study, we will try to select recommendations
so as to maximize the facet diversity. Consequently, we need to
evaluate the diversity in recommendation ranking. Evaluation of query
recommendations is also an important issue in this research area and
relatively less investigated than that of document search, as
evaluating diversified results is problematic even for this case.

\bibliographystyle{abbrv} 

\bibliography{my}

\end{document}